\documentclass[10pt,twocolumn,letterpaper]{article}

\usepackage{cvpr}
\usepackage{times}
\usepackage{epsfig}
\usepackage{graphicx}
\usepackage{amsmath}
\usepackage{amssymb}

\usepackage[pagebackref=true,breaklinks=true,letterpaper=true,colorlinks,bookmarks=false]{hyperref}

\cvprfinalcopy 

\newcommand{\degree}{\ensuremath{{}^{\circ}}\xspace}

\newcommand{\ignore}[1]{}

\def\etal{\emph{et al}\onedot}

\def\cvprPaperID{771} 

\ifcvprfinal\fi
\begin{document}

\title{Towards Automatic Image Editing: Learning to See another You}

\author{Amir Ghodrati$^{1}$\thanks{A. Ghodrati and X. Jia contributed equally to this work.},
Xu Jia$^{1}\footnotemark[1]$, Marco Pedersoli$^2$\thanks{LEAR project, Inria Grenoble Rhˆone-Alpes, LJK, CNRS, Univ. Grenoble Alpes, France.}, Tinne Tuytelaars$^1$ \\
$^{1}$KU Leuven, ESAT-PSI, iMinds \qquad  $^{2}$INRIA \\
$^{1}${\tt\small firstname.lastname@esat.kuleuven.be} \quad $^{2}${\tt\small marco.pedersoli@inria.fr}\\
}

\maketitle

\begin{abstract}
Learning the distribution of images in order to generate new samples 
is a challenging task due to the high dimensionality of the data and 
the highly non-linear relations that are involved.
Nevertheless, some promising results have been reported in the literature recently,
building on deep network architectures.
In this work, we zoom in on a specific type of image generation: 
given an image and knowing the category of objects it belongs to (e.g. faces), 
our goal is to generate a similar and plausible image, but with some altered attributes.
This is particularly challenging, as the model needs to learn to disentangle
the effect of each attribute and to apply a desired attribute change to a given input image, while keeping the other attributes and overall object appearance intact.
To this end, we learn a convolutional network, where the desired attribute information
is encoded then merged with the encoded image at feature map level.
We show promising results, both qualitatively as well as quantitatively, in the context of a retrieval experiment, on two face datasets (MultiPie and CAS-PEAL-R1).
\end{abstract}

\section{Introduction}
\label{sec:intro}


When looking at pictures of family and friends, does any of the following comments sound familiar to you? 
{\em``I took five pictures of this group, but none of them has everyone looking at the camera at the same time!''}
or {\em``Nice picture, just a pity her eyes are hidden behind those big sunglasses.''} Or maybe you ever wondered:
{\em``I look so old on this picture - wish I could make myself look 5 years younger!''} or {\em``What would he look like if he grew a beard?''}.
Now imagine the next generation of image editing tools, which will have learned to generate images of faces, so they can not just correct for red eyes, but also modify specific attributes of the people depicted in photographs: make them face the camera, remove their sunglasses, make them look younger or add that beard, while keeping all other facial characteristics
and imaging conditions constant.

\begin{figure}[t]
\centering
\includegraphics[width=1.0\linewidth]{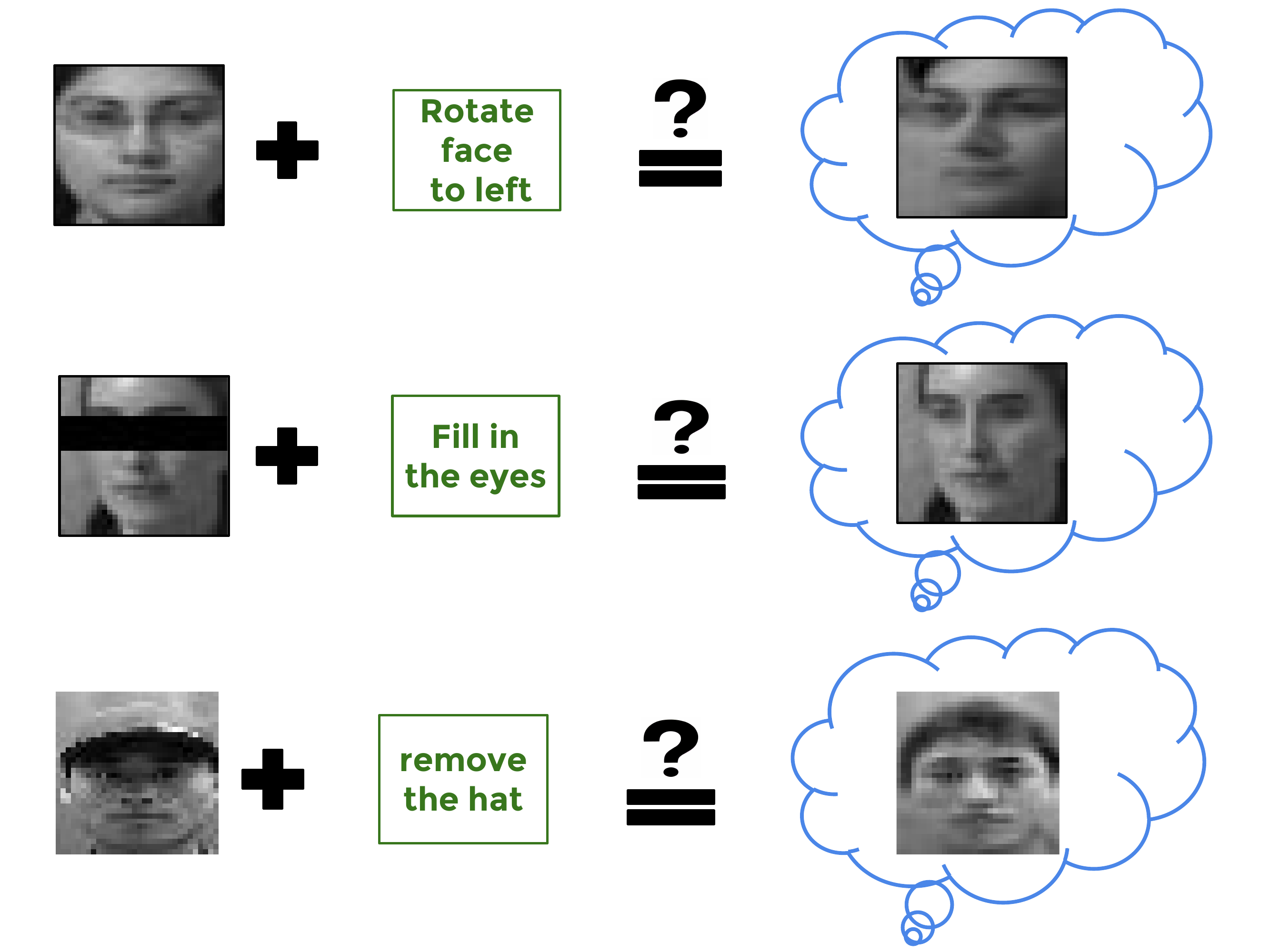} 
\caption{Generation of images in 3 different tasks using our proposed method.}
\label{fig:page1}
\end{figure}

Thoughts along these lines motivated us to start looking into this problem.
While our current method cannot yet be applied ‘in the wild’ in applications like the one sketched above, it does show this might actually become doable in the near future. 
What we can do now is shown in Figure~\ref{fig:page1}.

There have been several works on image generation~\cite{Dosovitskiy-cvpr15, Denton-nips15, Gauthier-14, Goodfellow-nips14, Hinton-icann11, Kingma-iclr14, Kulkarni-nips15, Li-icml15, Rezende-icml14, Tieleman-Thesis, Yang-nips15, Yim-cvpr15, zhu2013deep, zhu-nips14}. 
However, note how our problem setting is different from the one tackled by most image generation methods found in the literature. 
\cite{Denton-nips15, Gauthier-14, Goodfellow-nips14, Li-icml15} train a model to generate images from scratch, i.e. without an input image as reference.
Others often focus only on changing a face or object’s pose (e.g.~\cite{Dosovitskiy-cvpr15,Yang-nips15}), or learn to generate images for faces only for a canonical setting (e.g. looking straight into the camera, with standard diffuse illumination and neutral expression \cite{Yim-cvpr15,zhu2013deep}). 

Here is what we do.
We start with a large dataset of cropped and aligned faces, with a variety of attributes (e.g. pose, illumination, attributes like wearing-glasses or wearing-hat) annotated and varied systematically for each individual in the database.
We train a model that takes an face image as input and encodes the appearance of the face; takes a desired attribute vector as input and encodes it; then combines these two pieces of information and fuses them in a way that finally allows it to generate a new image depicting the same individual, yet with this one attribute changed. 

We propose to address this problem based on a convolutional encoder-decoder framework. 
The most critical part then is how to encode the desired attribute modification into a representation that is compatible with the input face image. 
Here we propose to encode the desired attribute into several feature maps which are compatible with the feature maps computed from the input face image. 
The two sets of feature maps are then integrated using a feature map fusion module. 
This is followed by a deconvolution module to generate the target face image.
In addition, in order to generate more realistic images we adopt the global-local two stage scheme, dividing the problems in two stages with each one focusing on one aspect. The first stage is in charge of rendering a global representation of the desired object, while the second focuses on local refinements to remove some artefacts.

To demonstrate the effectiveness of the proposed method on the above task, 
we evaluate it on three datasets for three tasks, that is, MultiPIE~\cite{multipie} for rotating faces, CAS-PEAL-R1~\cite{caspeal} for adding facial accessories and also a synthetic dataset based on MultiPIE for image completion.
The qualitative results show that the proposed method can generate face images of good quality and keep all the input face information except what is specified by the desired attribute.
For the experiment on the MultiPIE dataset, we also apply the proposed method to the task of face retrieval and quantitatively evaluate its performance.
Given a query image we want to retrieve similar images but with some altered attributes, similar to what is done in~\cite{Ghodrati-icmr15}. 
For instance we want to retrieve the same person but with a beard or sun glasses. 
Instead of learning an attribute detector first, we generate with our method an image with the desired attribute and then take the altered image as query for standard similarity-based retrieval.

In summary, the main contributions of this paper are: i) definition of a new problem, where the goal
is to generate images similar to a source image yet with one aspect changed;
ii) a solution to this problem, 
in the form of a new encoder-decoder based scheme for image generation, where the desired attribute
modification is first encoded then integrated at feature map level;
iii) the insight that the final result can be refined by adding a second stage CNN ; 
and iv) good results, both qualitatively as well as quantitatively, as evaluated in an image retrieval setting.

The remainder of this paper is organised as follows. 
First, in section~\ref{sec:related}, we discuss related work. 
Section~\ref{sec:method} describes our proposed method.
Section~\ref{sec:expres} details the experimental evaluation of our method,
and section~\ref{sec:conclu} concludes the paper. 

\vspace{-1mm}
\section{Related Work}
\label{sec:related}
Recently, there have been several works addressing the task of image generation
~\cite{Dosovitskiy-cvpr15, Denton-nips15, Gauthier-14, Goodfellow-nips14, Hinton-icann11, Kingma-iclr14, Kulkarni-nips15, Li-icml15, Rezende-icml14, Tieleman-Thesis, Yang-nips15, Yim-cvpr15, zhu2013deep, zhu-nips14}.
These methods can roughly be divided into two categories.

\begin{figure*}[t]
\begin{center}
    \includegraphics[width=1\linewidth]{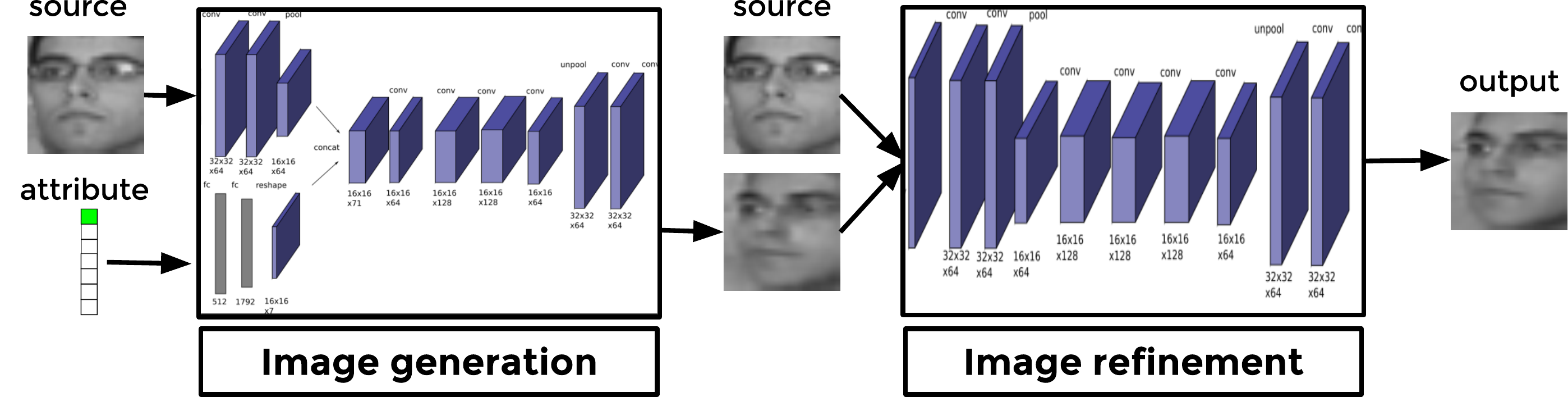}
\end{center}
\caption{An overview of our proposed method. Given a source image and attribute vector, we modify the source based on the attribute vector and generate an image in a two stage approach.}
\label{fig:overview}
\end{figure*}

First, several works follow an unsupervised approach.
They learn the distribution of natural images and generate an image from scratch~\cite{Denton-nips15, Gauthier-14, Goodfellow-nips14, Li-icml15}.
Other methods belonging to this category follow the encoder-decoder framework.
Hinton~\cite{Hinton-icann11} and Tieleman~\cite{Tieleman-Thesis} proposed the capsule network 
which uses the composite units in the code layer of an autoencoder to disentangle different visual components.
Kingma and Welling~\cite{Kingma-iclr14} and Rezende~\etal~\cite{Rezende-icml14} proposed the variation autoencoder 
which maximizes a lower bound of the likelihood. 
It applies the reparameterization trick to the latent variables in the code layer to model different factors of the visual appearance.

The second group of works follow a supervised approach.
Kulkarni~\etal~\cite{Kulkarni-nips15} proposed to use convolutional variational autoencoder and a special training scheme to learn interpretable graphics code for different appearance transformations.
Dosovitskiy~\etal~\cite{Dosovitskiy-cvpr15} map the parameter space which describes high-level information 
such as chair type, viewpoint and affine transformation into the 2D image space and generate different types of chairs.
Zhu~\etal~\cite{zhu2013deep,zhu-nips14}, Yim~\etal~\cite{Yim-cvpr15} and Yang~\etal~\cite{Yang-nips15} 
propose to generate images in a weakly-supervised manner.
Given an object in a specific pose, ~\cite{zhu2013deep} propose a network to generate frontal faces with fixed illumination. 
Yim~\etal~\cite{Yim-cvpr15} propose a network to generate a face with desired pose by coding the pose vector in the border of the image.
Yang~\etal~\cite{Yang-nips15} use a recurrent network to model pose changes in faces. 
Similarly, our method also falls in this category and focuses on the generation of face image. However in contrast to others, our method is not designed specifically for changing the pose of a face but we will show the generality of our network with different attributes like sunglasses and hats. In addition, most of the other methods modify other aspects of a face (\eg illumination) in order to generate a new exemplar that is more suitable for other tasks, such as recognition. In contrast in our task the challenge is to generate an image with the desired attribute while preserving the rest of image information as before.

Our work is also related to face image editing such as novel facial image generation.
In~\cite{Mohammed-siggraph09}, Mohammed~\etal proposed a two stage method for novel facial image generation.
They first make an estimation using a parametric global model and then use a local non-parametric model to refine the generation.
In our paper, we propose a pipeline which consists of an image generation network and an image refinement network
and show that it shares the same global-local two stage philosophy.

\vspace{-1mm}
\section{Proposed Method}
\label{sec:method}
In this section, we describe the proposed convolutional encoder-decoder architecture for image generation. The overview of our proposed method is shown in Figure~\ref{fig:overview}.
Given a source image and a target attribute vector, our goal is to generate a new image with the target attribute while maintaining as much as possible the appearance of the source image. To this end, we propose a two-stage approach, with a first network for image generation (see~\ref{subsec:stage1}) and a second one for image refinement(see~\ref{subsec:stage2}).
Let us denote the source image and source attribute as $\mathbf{X}$ and $\mathbf{C_X}$, and target image and target attribute as $\mathbf{Y}$ and $\mathbf{C_Y}$. 

\subsection{Convolutional Encoder-Decoder Architecture for Image Generation}
\label{subsec:stage1}
\begin{figure*}[t!]
\centering
\scalebox{0.8}
{
\includegraphics[width=1.0\textwidth]{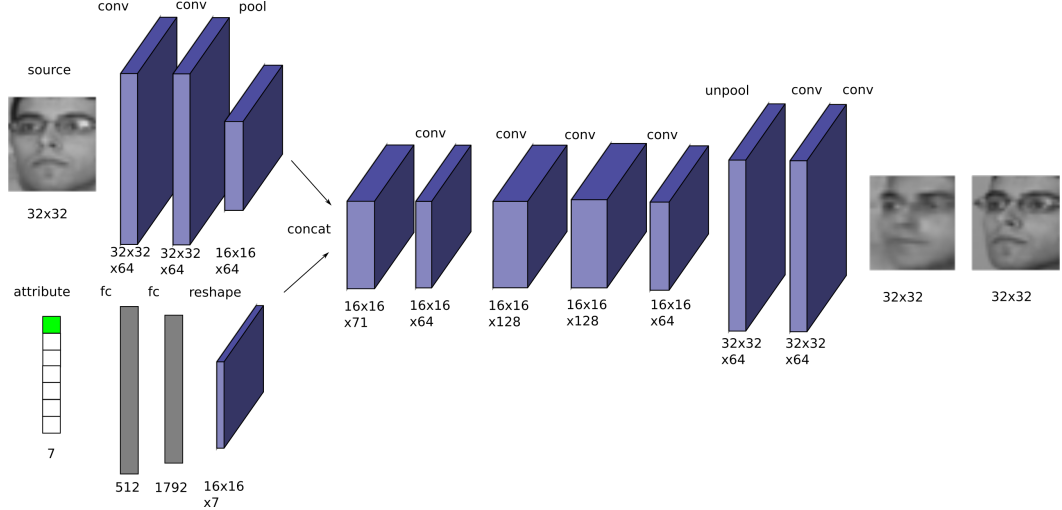}
}
\caption{Convolutional encoder-decoder architecture for image generation.}
\label{fig:network_stage1}
\end{figure*}

Inspired by the recent success of deconvolutional networks in generating accurate images of chairs 
from high-level descriptions~\cite{Dosovitskiy-cvpr15} and semantic segmentation~\cite{Long-cvpr15,Noh-iccv15}, 
we adopt a convolutional encoder-decoder architecture for our task. 
However, in contrast to previous works we deal with two inputs coming from different modalities: 
an image and a target attribute vector.
The network structure for image generation is shown in Figure~\ref{fig:network_stage1}.
The architecture can be conceptually divided into four operations:
image encoding, attribute vector encoding, feature map fusion and image decoding.

{\flushleft{\textbf{Image encoding.} }}
In this component, we encode a source image $\mathbf{X}$ into a set of feature maps via two convolutional layers and a max-pooling layer.
Inspired by the OxfordNet~\cite{Simonyan-iclr15}
we use two consecutive convolutional layers, each of which includes filters of size ($3\times3$) 
and one rectifier linear unit (ReLU)~\cite{Krizhevsky-nips12}.
Theese two of consecutive convolutional layers help to increase the receptive field without increasing the number of parameters too much.
The convolutional layers are able to capture local information of the source image 
and later we will show how to use them to generate the target image.
The structure of this part can be expressed as 
Conv($3,3,64$)-ReLU-Conv($3,3,64$)-ReLU-Pool($2,2$). 
Since we use a $32\times32$ image as input, we get feature maps of size $16\times16\times64$ at the end.

\begin{figure}[t]
\centering
\includegraphics[width=1.0\linewidth]{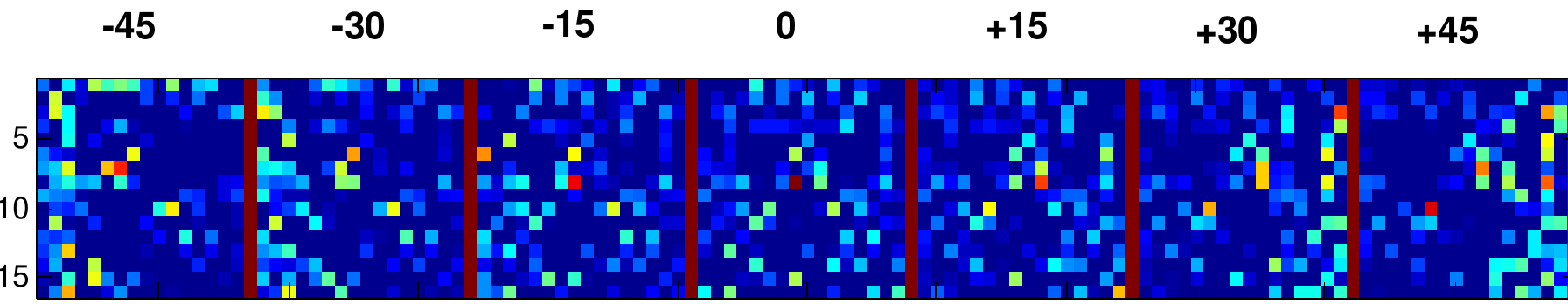} 
\caption{Attribute's feature maps (each of them $16\times16$). Here we have trained the network for different face poses as attributes. Depending on the input pose vector, responses are  higher in different parts of the feature maps and somehow make face-like shapes in different poses. Note that they are independent from the input image.}
\label{fig:feature_maps}
\end{figure}

{\flushleft{\textbf{Attribute vector encoding.} }}
We express the attribute as a one-hot vector and embed it into a higher dimensional space using a fully connected layer.
To make it fit to the feature map fusion step, 
we add another fully connected layer and then reshape it to several feature maps of size $16\times16$.
The reshaped feature maps have an approximate shape of the input image but with desired attributes.
The structure of this part can be expressed as
FC($512$)-FC($1792$)-Reshape($16,16,7$).
The output of this step is a set of feature maps of size $16\times16\times7$.
For illustration, we train a network to modify the pose of the input face with just a single $16\times16$ feature map and in Figure~\ref{fig:feature_maps} we visualize this feature map output from the attribute vector encoding step for different target poses. Note that these feature maps are independent from the source image but still make a rough estimation about shape of the face in different poses.
%
%
{\flushleft{\textbf{Feature map fusion.} }}
In this step, we fuse the feature maps obtained from the encoding of the input image and the attribute vector. 
The feature maps with attribute information can propagate their message to the feature maps extracted from the source input.
The two sets of feature maps are first stacked together, 
generating a new set of feature maps of $16\times16\times71$.
Then a feature map fusion layer similar to the cross channel pooling layer in~\cite{Lin-iclr14, Szegedy-cvpr15} 
is applied on top to fuse the two set of feature maps.
The cross channel pooling layer can be viewed as a ($1,1$) convolutional layer followed by the ReLU,
In our case, the feature map fusion layer is more sophisticated, 
including several consecutive convolutional layers.
The convolution operations compute the weighted average of feature maps
so that it allows sufficient interaction 
between feature maps of the source image and the attributes in a learning framework.
The structure of this part can be expressed as 
Concat-Conv($3,3,64$)-ReLU-Conv($3,3,128$)-ReLU-Conv($3,3,128$)-ReLU-Conv($3,3,64$)-ReLU.
The output of this step is a feature map of size $16\times16\times64$.

{\flushleft{\textbf{Image decoding.} }}
Once the feature maps computed from source image $\mathbf{X}$ and the attribute vector $\mathbf{C_X}$ have been fully integrated, 
the next step is to generate an image with the same size as the input image.
This module aggregates local information from different feature maps.
Similar to~\cite{Dosovitskiy-cvpr15}, 
a deconvolution module here consists of a $2\times2$ unpooling layer and two convolutional layers.
The unpooling layer doubles the size of feature maps in the previous layer 
by replacing each element of the feature map with a $2\times2$ block.
The top left corner is filled in with the element of the feature map and the rest are filled with zeros.
The unpooling layer is followed by two consecutive ($3,3$) convolutional layers.
Note that all convolutional layers have a ReLU activation except the last one 
because the output should have both positive and negative values after being normalized to zero mean and one standard deviation.
The structure of this part can be expressed as 
Unpool-Conv($3,3,64$)-ReLU-Conv($3,3,1$).
The output of this step is an image of size $32\times32\times1$.

\subsection{Image Generation Refinement}
\label{subsec:stage2}
The ideal generated image should have good quality, have the desired target attributes and keep the appearance of the input image.
It is difficult for a single network to generate an image that satisfies all the above requirements simultaneously.
Therefore, we adopt the divide-and-conquer scheme, dividing the problem into two stages with each one focusing on a different aspect.
The output of the proposed convolutional encoder-decoder network is already a reasonable generation but with some details missing and some artifacts.
To reduce the load of the network, we propose adding another convolutional encoder-decoder network to perform image refinement in a second stage.
The second stage takes as input the source image and the generated image of the first stage. 
These two inputs are first concatenated channel-wise then we apply several convolutional, ReLU and max-pooling layers in the encoding process followed by unpooling, convolutional and ReLU layers in the decoding process.
Convolutoinal layers locally fuse the information from two inputs and refine the output of the first stage network.
The use of multiple consecutive convolutional and max-pooling layers enlarges the receptive field,
making a big displacement for some pixels possible.
The architecture of the whole network is as follows: Concat-Conv($3,3,64$)-Conv($3,3,64$)-Pool($2,2$)-Conv($3,3,128$)-Conv($3,3,128$)-Conv($3,3,128$)-Conv($3,3,64$)-Unpool($2,2$)-Conv($3,3,64$)-Conv($3,3,1$).

\subsection{Relationship to Global-Local Two Stage Method}
We show that our method can be viewed as a global-local two stage method 
used for synthesizing novel faces~\cite{Mohammed-siggraph09}.
The traditional global-local two stage method first learns a global parametric model 
to make an approximate estimation of the global structure of a face.
Then a local non-parametric model is conditioned on the global estimation to reproduce realistic local texture.
In~\cite{Mohammed-siggraph09}, an image is divided into several overlapped regions
and for each region a separate library of patches is stored.
Then, the initial result of the global model is refined using the stored patch library.

We can view our full pipeline in a global-local perspective as well: The image generation network generates an initial image with our desired global changes and the refinement network locally refines the output of the first stage using convolutional filters. In the first stage, the goal is to generate an image containing the correct global structure of our target face. In the second stage, the goal is to keep the visual consistency with the outputs of the first stage but removing artifacts and adding the details to the face. We show the effectiveness of such paradigm for image generation in the section~\ref{sec:expres}.

\subsection{Training}
We train the two networks successively.
The input of the first network are the source image of size $32\times32$ and a one-hot vector.
Each image is preprocessed by subtracting the mean and dividing by the standard deviation.
Then we feed the output of the first network and the source image as input to the second network.
For both networks, we use Mean Squared Error (MSE) as the loss function,
\begin{eqnarray}
 & L_1(W_1) = || F_1(X, C_X, W_1) - Y ||_2^2, 
 \\
 & L_2(W_2) = || F_2(X, \hat{Y_1}, W_2) - Y ||_2^2,
\end{eqnarray}

The training parameters are fixed for both networks.
The training is carried out using mini-batch gradient descent with backpropagation~\cite{LeCun-89}
and the batch size is set to 32.
We use a fixed momentum of 0.95 and learning rate of $1e-5$.
All the weights are initialized with the method proposed in~\cite{He-iccv15}.

\vspace{-1mm}
\section{Experiments}
\label{sec:expres}
In order to demonstrate the effectiveness of the proposed method we evaluate it for three different tasks.
The main task is to rotate the face and is carried out on MultiPIE dataset~\cite{multipie}.
We extensively evaluate our method for this task, showing both qualitative and quantitative results.
The other two tasks are generating faces with different accessories (sunglasses and hats) on the CAS-PEAL-R1~\cite{caspeal} dataset and filling in the missing part for a face image on synthetic data generated from MultiPIE. 
We show some qualitative results for these two tasks.

\subsection{Rotating Faces}
\label{subsec:multipie}
The session 1 of MultiPIE dataset consists of images of 249 identities under 15 poses and 20 different illumination conditions. We select a subset of it that covers 7 poses ($-45\degree$ to $+45\degree$). The first 100 subjects are used for training and the rest are used for testing. 
All faces are aligned and cropped based on eyes and chin annotations provided by Shafey \etal~\cite{Shafey-pami13}, then resized to $32\times32$ pixels and converted to grayscale.

The goal of the method is to preserve as much as possible all the aspects of the input image and just change the target attribute. Here we consider pose as an attribute and aim at generating faces with the same identity and illumination as the input image but with the desired pose. The input to our method is an image and a target pose vector. 
To this end, during training we build a set of image pairs, where the two image in a pair show the same person, with the same illumination, but with different pose (i.e. $100\times20\times P_{7}^{2}=84000$ pairs). The first element of a pair is considered as input image of the network and the second element is the ground-truth target image.

\subsubsection{Image Generation Results}
\label{subsubsec:results}
\begin{figure}[t]
\begin{center}
   \includegraphics[width=0.8\linewidth]{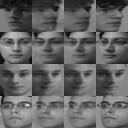}
\end{center}
\caption{Some qualitative results of our image generation from test data of MultiPIE. In each row, first column is input image, last column is ground-truth target image, 2nd column is output of first stage and 3rd column is generated image of second stage network.}
\label{fig:2stage_examples}
\end{figure}

In Figure~\ref{fig:2stage_examples} we show some qualitative results, i.e., the generated images from both stages of the method. We can see the proposed method can generate face images that are visually similar to the target face, no matter identity, pose or illumination.
Notice how the generated face images from the second stage have better image quality and details than first stage outputs.

We validate the effectiveness of the second stage network quantitatively as well. We randomly select $10000$ generated images and compare the performance of each stage in terms of per-pixel mean squared error (MSE) between generation and ground-truth image. Following the ``mean per-pixel reconstruction error'' term in~\cite{turmukhambetov2015modeling}, we call this metric ``mean per-pixel generation error'' and report it in Table~\ref{table:2stage_MSE_performance}. 
The generation of the second stage is better than the first stage both qualitatively and quantitatively.
This implies that the second stage network works as expected: it locally refines the initial generation via convolutional layers.

\begin{table}
\begin{center}
{\small
{
\begin{tabular}{|c|c|}
\hline
stage & MSE \\
\hline\hline
{first stage}	&	$380.89$ \\
{second stage}	&	$369.16$ \\
\hline
\end{tabular}
}
}
\end{center}
\caption{Mean Per-Pixel Generation Error (Mean Squared Error) on randomly selected 10000 test pairs.}
\label{table:2stage_MSE_performance}
\end{table}

\subsubsection{Quantitative Results on Retrieval Task}
\label{subsubsec:retrieval}
To demonstrate how our method preserves other aspects of a face (like identity and illumination) while changing its pose, we apply it to the task of face retrieval and quantitatively evaluate its performance. 
We randomly select 10000 query images and target poses (same as last subsection) from test data and retrieve the top K similar images from test set (containing all the $20860$ faces). 
We first generate an image with the target pose (or extract intermediate features) using the proposed method and then retrieve the K most similar faces to the generated face. However, we find there is a different distribution between the generated images and the original images from the dataset. To address this difference, we generate the 7 possible poses for each original face and keep the one that is closest to the original one (using pixel level Euclidean distance). With this procedure, images of both the query and the candidate pool have a similar distribution. This process can be done offline, just once, without any extra supervision. 

We compare our method with several intuitive 2-step retrieval baselines. 
In the first step we train a pose classifier to filter out all faces except those with the same pose as the target face and in the second step we rank the remaining faces by comparing them with the query using different features. 
For the first step, we train the pose classifier using features extracted from ``fc6'' layer of~\cite{Parkhi-bmvc15} and use it for all the experiments in 2-step retrieval scenario since it outperforms other features like ``fc7'' of~\cite{Simonyan-iclr15} and ``fc7'' of~\cite{Parkhi-bmvc15} in pose estimation. The pose classifier obtains a competitive $73.3\%$ accuracy on test data.
In all experiments we L2-normalized raw pixel images or features and compare them using Euclidean distance. 
We have two criteria for evaluation.
For the first one, we consider a retrieved image as correct if its identity is the same as the input and its pose is as the target input pose. 
For the second criterion, we further require that the correct image should also have the same illumination as the input face. 
We report our results using recall@1, recall@5 and recall@20 metrics. recall@K is defined as the percentage of queries for which at least one correct image is among the top K retrieved images. 

In table~\ref{table:retrieval_performance_2} we report the performance of our method and the baseline for the first criterion. For our method we use feature maps extracted from the middle layer of the second stage network. 
For VGG-16~\cite{Simonyan-iclr15} and Deep features~\cite{Parkhi-bmvc15} we extract features from ``fc7'' and ``fc6'' layers respectively. Note that both the other two networks have 16 layers and are trained with millions of images. 
Table~\ref{table:retrieval_performance_3} shows the retrieval performance with the second criterion. In this case our method obviously outperforms the other baselines. 
To be a fair comparison, in this table we report the results just for those features that illumination information is coded in. Deep and VGG-16 features are trained to be robust against illumination changes so they have poor results in this criterion while we intentionally want to keep all the input face information except what is specified by the input attribute vector.

\begin{table}
\begin{center}
{\small
{
\begin{tabular}{|c|c|c|c|c|}
\hline
Method & Features & R@1 & R@5 & R@20\\
\hline\hline
{2-step retrieval}	& image		& 11.5\% & 25.9\% & 43.2\% \\
{2-step retrieval}	& our 2nd network	& 16.4\% & 33.0\% & 50.7\% \\
{2-step retrieval}	& VGG-16	& 21.7\% & 38.1\% & 53.5\% \\
{2-step retrieval}	& Deep		& 56.9\% & 78.0\% & 90.2\% \\
\hline
Yim~\etal~\cite{Yim-cvpr15} & CPI 	& \bf{61.0}\% & 79.5\% & \bf{90.8}\% \\
\hline
{ours}			& image		& 39.0\% & 61.5\% & 75.4\% \\
{ours}			& our 2nd network	& 53.5\% & \bf{80.5}\% & 90.7\% \\
\hline
\end{tabular}
}
}
\end{center}
\caption{The retrieval performance for the first criterion. A retrieved image is considered as correct if its identity and pose are correct.}
\label{table:retrieval_performance_2}
\end{table}

\begin{table}
\begin{center}
{\small
{
\begin{tabular}{|c|c|c|c|c|}
\hline
Method & Features & R@1 & R@5 & R@20\\
\hline\hline
{2-step retrieval}	& image		& 6.3\% & 15.6\% & 25.3\% \\
{2-step retrieval}	& our 2nd network	& 9.5\% & 21.5\% & 32.7\% \\
\hline
Yim~\etal~\cite{Yim-cvpr15} & CPI 	& 14.8\% & 39.7\% & 63.7\% \\
\hline
{ours}			& image		& 29.5\% & 52.8\% & 66.3\% \\
{ours}			& our 2nd network	& \bf{43.2}\% & \bf{74.1}\% & \bf{86.0}\% \\
\hline
\end{tabular}
}
}
\end{center}
\caption{The retrieval performance for the second criterion. A retrieved image is considered as correct if its identity, pose and illumination are correct.}
\label{table:retrieval_performance_3}
\end{table}

We also compare our method with~\cite{Yim-cvpr15} in retrieval setting. 
To this end authors kindly sent us the CPI features they have used for face recognition task. 
As shown in table~\ref{table:retrieval_performance_2} they outperform in recall@1 but we are better in recall@5. 
However, they explicitly train the network for just keeping the identity which is the criterion we use in table~\ref{table:retrieval_performance_2}.
According to the results in table~\ref{table:retrieval_performance_3} which is more interesting for our goal
we are outperforming them in all cases. 
Note that though they do not explicitly model illumination, this information is coded in their CPI features.

\begin{figure}[t]
\begin{center}
   \includegraphics[width=1\linewidth]{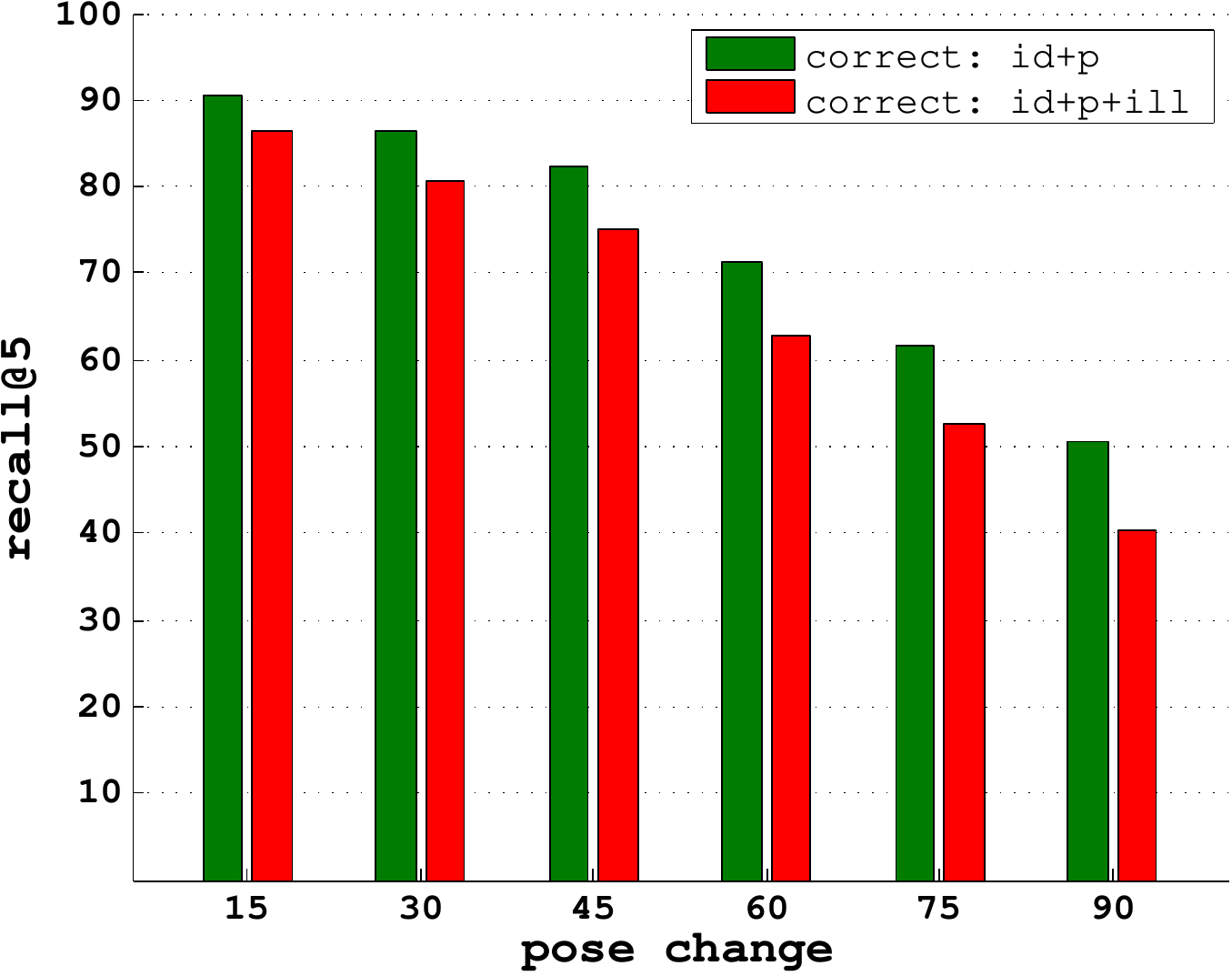}
\end{center}
\caption{Analysis of retrieval results on various pose changes (from input pose to target pose).}
\label{fig:pose_change}
\end{figure}

\begin{figure}[t]
\centering
\includegraphics[width=1.0\linewidth]{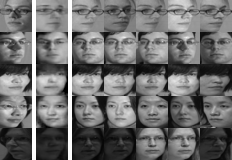} 
\caption{Some visual retrieval results using our method. First column is query, second one is generated image using our method and the rest columns are 5 top similar faces to generated face in feature space.}
\label{fig:retrieval_results}
\end{figure}

To get some insight about the performance of the generated images under various pose changes, we divide the test data into 6 groups in terms of the pose change from input pose to desired pose. Figure~\ref{fig:pose_change} shows the retrieval performance (recall@5) for each group separately. As could be expected, small pose changes are easier to deal with than large ones, with the recall@5 gradually decreasing.
Some qualitative results of retrieval by generation are also shown in Figure~\ref{fig:retrieval_results}.

\subsubsection{Comparison with state-of-the-art}
\label{subsubsec:soa}

\begin{figure*}[t]
\begin{center}
   \includegraphics[width=1\linewidth]{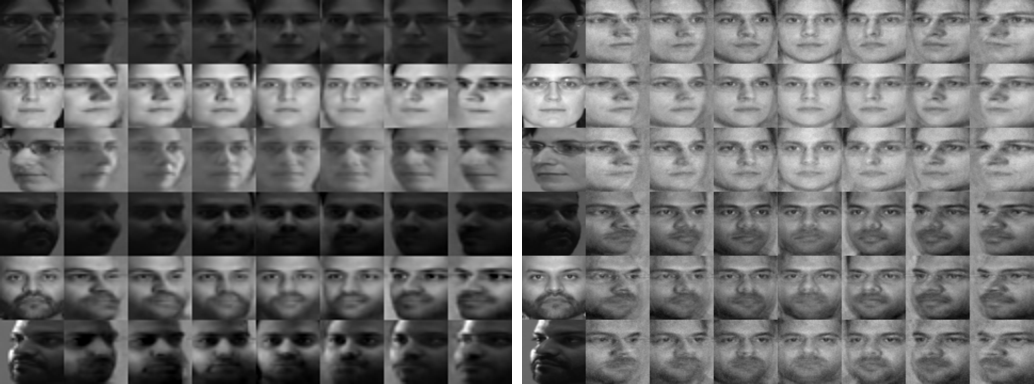}
\end{center}
\caption{Qualitative comparison between our generated images (Left) and~\cite{Yim-cvpr15} (Right). On each set the first column is the input face and next 7 ones are generated faces in different poses.}
\label{fig:qualitative_compare}
\end{figure*}

We show some images generated by our method and the method of~\cite{Yim-cvpr15} in Figure~\ref{fig:qualitative_compare}. 
From the comparison between our method and theirs, we can see that our method preserves the source image information better than theirs even though their images are a little sharper than ours. 
One possible explanation is that they trained their network on $60\times60$ pixel images while we trained on $32\times32$ pixel images. 
It is worth mentioning that their method is much more complex than ours.
Due to the use of locally linear and fully connected layers, their network has many more parameters than ours. 
Besides, they add an auxiliary branch for reconstruction in their network to preserve the identity.
There are only two convolutional encoder-decoder networks in our method and it can generate face images that preserve the identity better than theirs based on the qualitative results.
In addition, they train the network to generate neutral illumination regardless of the input image illumination.

\subsection{Adding Facial Accessories}
\label{subsec:caspeal}
\begin{figure}[t]
\begin{center}
{
  \includegraphics[width=1\linewidth]{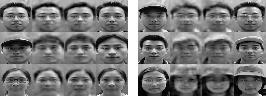}
}
\end{center}
\caption{Some qualitative results of our image generation from test data. Each row contains 2 set of examples. In each example, the first column is the input image, the last column is the desired image, 2nd column is generation of first stage network and 3rd column is the generated image of the second stage network.}
\label{fig:caspeal}
\end{figure}

In this experiment, we use images with 7 attributes including 3 different hats, 3 different glasses and a face without any accessory. To this end, we use CAS-PEAL-R1~\cite{caspeal} dataset and separate the data based on person identities and use the first 350 identities for training and the other 88 identities for test. All faces are aligned and cropped based on eyes and then resized to $32\times32$ pixels and changed to grayscale image.
We use the same strategy as for the MultiPIE dataset to make a pair list of all possible permutations of attribute changes (in total $2442$) in order to train our networks. Figure~\ref{fig:caspeal} shows some qualitative results of our method for test data.
Note that our method can remove one accessory (e.g. hat) and also substitute it with another attribute (e.g. glasses) at the same time. 
As is shown in the Figure~\ref{fig:caspeal}, the second stage of our method (3rd column) refines the generation and creates better quality images compared to our first stage (2nd column).
%
One explanation is that different people have different ways of wearing hat and glasses and people are not under the same condition when wearing different accessories. 
Such variations increase uncertainty of the data and lead the network to predict the average image which in practice results in blurry predictions.
Also, the alignment is only carried out based on the eyes annotation provided in the dataset, and it is quite approximative for some samples.
Instead, for MultiPIE, the images are recorded at the same time with cameras from different views and the alignment is carried out based on both eyes and chin.
This explains why for the experiment on MultiPIE dataset we have better alignment and hence better image generation quality than the experiment on CAS-PEAL-R1 dataset.


\subsection{Image Completion}
\label{subsec:inpainting}
\begin{figure}[t]
\centering
\scalebox{0.7}
{
\includegraphics[width=1.0\linewidth]{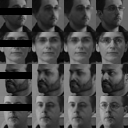} 
}
\caption{Filling in the region for eyes by our method. First column is input image, last column is desired target image, 2nd column is generation from first stage network and 3rd column is generated image of second stage network.}
\label{fig:synthetic_results}
\end{figure}

We carry out another interesting experiment on MultiPIE dataset to demonstrate the ability of the proposed method in generating images.
In this setting, we manually add a black bar to occlude the region of the eyes of a face image.
Then we train the proposed model to learn to generate a face image with eyes.
Note that to make it more robust to the influence of uncertainty behind the black bar, we use Mean Absolute Error instead of Mean Squared Error in this experiment.
Even though there are many ways that this region can be filled, our method can still generate reasonably good images as shown in Figure~\ref{fig:synthetic_results}.
The filled region is visually consistent with the non-occluded part of the face.
In addition, we can see that both networks work very well in this task.
The explanation for this is the task may be easy so that even the first network can generate an image with eyes and of good quality.

\vspace{-1mm}
\section{Conclusion}
\label{sec:conclu}
In this work we define a new problem, where the goal is to generate images with modified attributes while maintaining as much as possible the similarity to the original image.
We propose a solution to this problem, for the case of cropped and aligned faces, in the form of a two stage scheme with the first network for image generation and the second one for image refinement.
We have validated our approach for the case of faces, where we have shown that we can change the face pose, as well as other facial attributes while maintaining other aspects of the image. 
Our model can alter a face without human intervention and with noticeable accuracy. We have also shown that our model can be used for face pose retrieval with results comparable with state-of-the-art approaches.
For future work we would like to extend our method to address more challenging scenarios like dealing with misaligned input faces or applying it to different object categories.

{\small
\bibliographystyle{ieee}
\bibliography{egbib}
}

\end{document}